\documentclass[a4paper, superscriptaddress, amsfonts, amssymb, amsmath, reprint, showkeys, nofootinbib, twoside]{revtex4-1}
\usepackage[english]{babel}
\usepackage[utf8]{inputenc}
\usepackage{graphicx}
\usepackage{afterpage}
\usepackage[colorinlistoftodos, color=green!40, prependcaption]{todonotes}
\usepackage{amsthm}
\usepackage{mathtools}
\usepackage{physics}
\usepackage{xcolor}
\usepackage{graphicx}
\usepackage[left=23mm,right=13mm,top=35mm,columnsep=15pt]{geometry} 
\usepackage{adjustbox}
\usepackage{placeins}
\usepackage[T1]{fontenc}
\usepackage{lipsum}
\usepackage{csquotes}
\usepackage[pdftex, pdftitle={Article}, pdfauthor={Author}]{hyperref} 

\begin{document}
\title{Origami crawlers: exploring a single origami vertex for complex path navigation}

\author{Davood Farhadi}
    \email[Correspondence email address:]{dfarhadim@gmail.com}
    \affiliation{J.A. Paulson School of Engineering and Applied Sciences, Harvard University, Cambridge, MA 02138, USA}
    \affiliation{Department of Precision and Microsystem Engineering, Delft University of Technology, Mekelweg 2, 2628 CD Delft, The Netherlands}

\author{Laura Pernigoni}
    \affiliation{J.A. Paulson School of Engineering and Applied Sciences, Harvard University, Cambridge, MA 02138, USA}
    \affiliation{Department of Aerospace Engineering, Politecnico di Milano, Milan, 20133 Italy}

\author{David Melancon}
    \affiliation{J.A. Paulson School of Engineering and Applied Sciences, Harvard University, Cambridge, MA 02138, USA}
    
\author{Katia Bertoldi}
    \email[Correspondence email address:]{bertoldi@seas.harvard.edu}
    \affiliation{J.A. Paulson School of Engineering and Applied Sciences, Harvard University, Cambridge, MA 02138, USA}

\date{\today} 

\begin{abstract}
The ancient art of origami, traditionally used to transform simple sheets into intricate objects, also holds potential for diverse engineering applications, such as shape morphing and robotics. In this study, we demonstrate that one of the most basic origami structures—a rigid, foldable degree-four vertex—can be engineered to create a crawler capable of navigating complex paths using only a single input. Through a combination of experimental studies and modeling, we show that modifying the geometry of a degree-four vertex enables sheets to move either in a straight line or turn. Furthermore, we illustrate how leveraging the nonlinearities in folding allows the design of crawlers that can switch between moving straight and turning. Remarkably, these crawling modes can be controlled by adjusting the range of the folding angle’s actuation. Our study opens avenues for simple machines that can follow intricate trajectories with minimal actuation.
\end{abstract}


\maketitle

\section*{Introduction}
While traditionally regarded as an art form, origami has recently made a profound impact on engineering applications, playing a pivotal role in the design of deployable and reconfigurable structures \cite{hawkes2010programmable,zirbel2013accommodating, filipov2015origami, nelson2019origami, lee2021high, melancon2021multistable}, metamaterials with programmable responses \cite{silverberg2014using, schenk2013geometry} and mechanical logic elements \cite{treml2018origami, liu2023discriminative}. Further, origami principles have provided an opportunity to simplify and accelerate the fabrication of robots, since they enable their manufacturing from a flat composite through folding \cite{ma2013controlled, felton2013robot, felton2014method, rus2018design, liu2021micrometer, yan2023origami}. Beyond simplifying the manufacturing process, origami has also enabled the design of grippers to encapsulate delicate specimens \cite{teoh2018rotary} and efficient crawlers \cite{koh2012omega,onal2012origami,kim2019bioinspired,bhovad2021physical, ze2022soft}. 

In applications, crawlers are typically required to follow arbitrary trajectories. This is commonly accomplished by employing multiple actuators, each assigned to a specific gait, and coordinated through control circuits \cite{shepherd2011multigait, felton2014method, de2018inverted, wu2024modular}. The pursuit of simpler strategies for achieving locomotion along arbitrary trajectories remains a significant challenge in robotics \cite{yang2018grand}.
 In the case of soft machines, this challenge has been addressed by implementing actuation through a magnetic field \cite{hu2018small, ze2022spinning} and by integrating pneumatic control circuits \cite{lee2022buckling, drotman2021electronics}. Conversely, for rigid-body linkages, diverse gaits from a single actuation have been successfully achieved by incorporating passive actuators to modify the effective length of the linkage \cite{zarrouk20141star, noji2022modeling, feshbach2023curvequad}.

In this work, we leverage the highly nonlinear kinematics of origami to create a crawler capable of following complex trajectories with a single input. We focus on one of the simplest origami building blocks, the degree-four origami vertex \cite{abel2016rigid, waitukaitis2015origami, waitukaitis2016origami}, and demonstrate through a combination of experiments and modeling how the position of the central vertex and the orientation of the creases significantly impact its crawling ability and resulting trajectory. Guided by our model, we identify designs that can  move straight when the folding angle is actuated within a certain range and turn  when actuated in another range. This enables the creation of simple machines that can   follow complex trajectories by simply controlling the folding angle at the actuated crease.

\section*{Results}
\begin{figure}[t!]
    \centering
    \includegraphics[width=\linewidth]{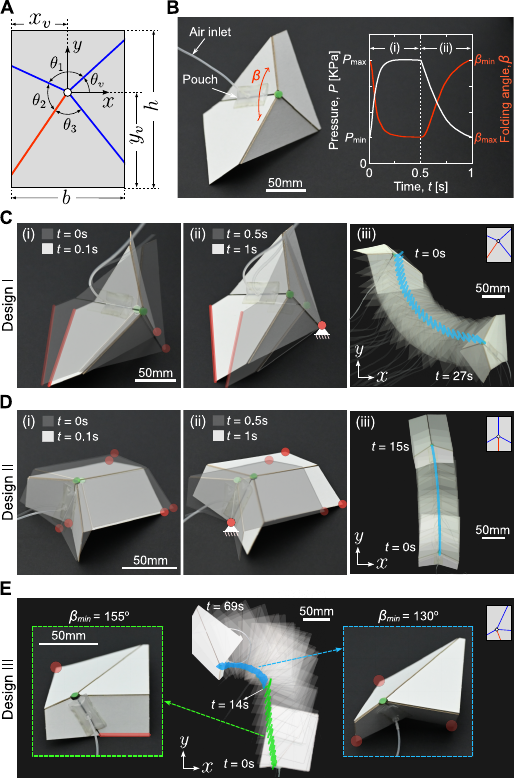}
    \caption{\textbf{Multimodal crawling locomotion of a degree-four origami vertex.} (A) Schematic of a degree-four origami vertex with mountain folds (blue lines), and a valley fold (red line). (B) Physical prototype actuated by pressurizing an inflatable pouch attached to the valley fold. The pressure profile used for actuation is shown on the right, together with the corresponding evolution of the folding angle $\beta$. (C)-(E) Experimental results for (C) Design I, (D) Design II, and (E) Design III. Snapshots of origami sheets at different times during loading cycles and frames from videos recorded during tests in which the samples were subjected to multiple inflation/deflation cycles. }
    \label{figure_1}
\end{figure}

We consider a rectangular sheet with width $b$ and height $h$ and embed four creases to create a generic degree-four vertex (Fig. \ref{figure_1}A). The flat state configuration of the sheet is defined by four sector angles, $\theta_i$ ($i=1,...,4$), the orientation of the creases with respect to the rectangular sheet, defined by the angle $\theta _v$, and the position of the central vertex, specified by the coordinates $(x_v, y_v)$.  Our focus is on developable (for which $\sum\theta_i=2\pi$) and rigidly foldable (for which.  $\theta_j < \sum \theta_{i\neq j}$) configurations \cite{abel2016rigid}. While these configurations may not always fulfill the flat foldability criterion (i.e., $\theta_1 + \theta_3 \neq \pi$), their kinematic is characterized by a single degree of freedom  (Movie S1).   To actuate the origami sheet, we attach an inflatable pouch to the valley crease (Fig. \ref{figure_1}B) and increase its pressure from $\text{P}_{\text{min}}$ to $\text{P}_{\text{max}}$ in 0.5 s.  This inflation step is immediately followed by a deflation phase, in which the pressure is reduced back to $\text{P}_{\text{min}}$ over 0.5 s.  This applied pressure profile induces an asymmetric rate of change in the folding angle of the valley crease, $\beta$. As shown in Fig.~\ref{figure_1}B, $\beta$ rapidly decreases from $\beta_{\max}$ to $\beta_{\min}$ within 0.1 seconds during inflation and returns more gradually to $\beta_{\max}$ during deflation. 

In Figs. \ref{figure_1}C and D, we show results for two different origami sheets, both with $b=90$ mm, $h=130$ mm, but different angles $\theta_i$ and $\theta_v$ as well as vertex coordinates. Specifically, Design I features an asymmetric pattern with angles ($\theta_1$, $\theta_2$, $\theta_3$, $\theta_v$)=$(\ 109^\circ, \ 79^\circ, \ 80^\circ, \ 45^\circ)$ and vertex coordinates $(x_v, y_v)=(45, \ 80)$ (Fig.~\ref{figure_1}C). In contrast, Design II, characterized by angles ($\theta_1$, $\theta_2$, $\theta_3$, $\theta_v$)=$(\ 110^\circ, \ 70^\circ, \ 70^\circ, \ 90^\circ)$ and vertex coordinates $(x_v, \ y_v)=(45, \ 50)$ mm, exhibits an axis of symmetry aligned with the long edge of the sheet (Fig.~\ref{figure_1}D).

\begin{figure*}[t!]
    \centering
    \includegraphics[width=\linewidth]{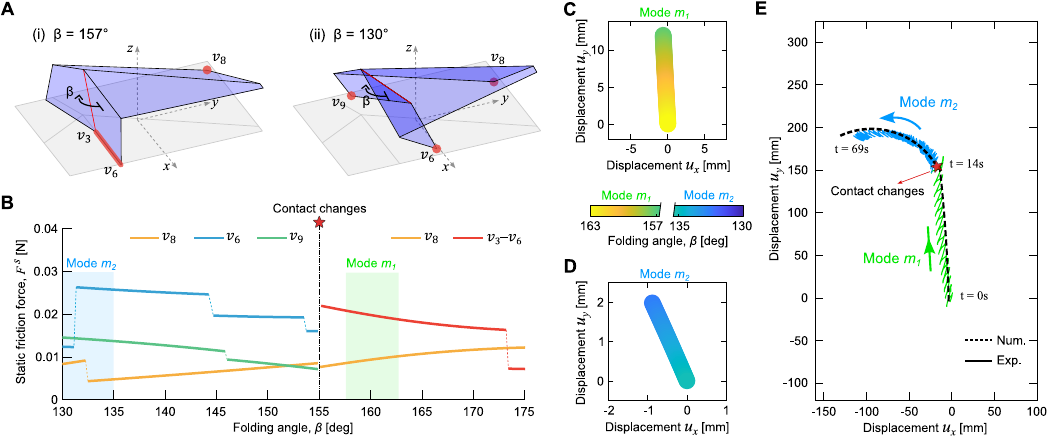}
    \caption{\textbf{Comparison of experimental and numerical results for Design III.} (A) Numerically predicted location of contacts (highlighted in red) for $\beta=157^\circ$ and $130^\circ$.  (B) Numerically predicted evolution of the static friction forces at the contact nodes as a function of folding angle $\beta$.  The ranges of $\beta$ used in the tests shown in Fig.~\ref{figure_1}E to move straight and turn are indicated by the green and blue shaded areas, respectively. (C)-(D) Numerically predicted displacement of the center of mass of the sheet for a complete actuation cycle with (C) $157^\circ\leq\beta\leq 163^\circ$ and (D) and $130^\circ\leq\beta\leq 135^\circ$. (E) Comparison of the numerically predicted (dashed line) and experimentally observed (continuous line) trajectory traced by Design III when subjected to 14 loading cycles with $157^\circ\leq\beta\leq 163^\circ$ followed by 55 loading cycles with $130^\circ\leq\beta\leq 135^\circ$.}
    \label{figure_2}
\end{figure*}

In Figs. \ref{figure_1}C(i)-(ii) and \ref{figure_1}D(i)-(ii) we show snapshots of the origami sheets at $t=0$, 0.1, 0.5 and 1 s, while in Figs. \ref{figure_1}C(iii) and \ref{figure_1}D(iii), we report frames recorded during tests in which we cyclically actuated the origami sheets. Four key observations can be made. Firstly, the orientation of the creases significantly influences the folding of the sheet, thereby determining the location of its center of mass. This, in turn, dictates the edges and vertices in contact with the substrate (highlighted by red lines and markers in Figs. \ref{figure_1}C(i) and \ref{figure_1}D(i)). 
Secondly, since the folding angle $\beta$ rapidly decreases during inflation (Fig.~\ref{figure_1}B),  all contact points slip due to the dynamic nature of the process (Figs. \ref{figure_1}C(i) and \ref{figure_1}D(i)). 
Thirdly, since the deflation process occurs at a comparatively slower rate (Fig.~\ref{figure_1}B), contact points with higher static friction forces tend to stick and act as anchor points (Figs. \ref{figure_1}C(ii) and \ref{figure_1}D(ii)). 
Fourthly, at the end of the loading cycle, both sheets move following the trajectory outlined by their anchor points (Figs.~\ref{figure_1}C(ii) and \ref{figure_1}D(ii)). Consequently, upon cyclic actuation, Design I crawls towards the left (Fig. \ref{figure_1}C(iii), and Movie S2), while Design II moves straight (Fig. \ref{figure_1}D(iii), and Movie S3).

The results of Figs.~\ref{figure_1}C and \ref{figure_1}D underscore the potential of exploiting the nonlinear kinematics of the origami sheets for locomotion, albeit with each design limited to a predefined trajectory. However, it is important to note that certain origami sheets offer opportunities for motion along arbitrary trajectories. As an example, in Fig. \ref{figure_1}E, we present results for an origami sheet (Design III) with  $b=90$ mm, $h=135$ mm, $(x_v, \ y_v)=(27, \ 36)$ mm, and  ($\theta_1$, $\theta_2$, $\theta_3$, $\theta_v$)=$(\ 140^\circ, \ 100^\circ, \ 60^\circ, \ 60^\circ)$. We find that for $157^\circ \leq \beta \leq 163^\circ$ this origami sheet moves forward upon cyclic actuation, while for $130^\circ \leq \beta \leq 135^\circ$ it turns left (see Movie S4). This change in locomotion is due to a shift in contact points. For $\beta>155 ^\circ$, an edge and a vertex of the sheet are in contact with the substrate (highlighted by red markers in Fig.~\ref{figure_1}E). However, when $\beta$ decreases below $155^\circ$, the shape change causes the origami device to establish new contact points at the three edges of the origami sheet (also highlighted by red markers in Fig.~\ref{figure_1}E). This shift in contact points changes the anchor point and, therefore, leads to a different crawling mode. 

Next, we develop a quasi-static model to systematically explore how the fold arrangement within the origami sheet affects its crawling trajectory. While an origami sheet itself is as a single-degree-of-freedom mechanism, its interactions with the substrate introduce three additional degrees of freedom: the components of displacement of its center of mass in the plane of the substrate, and the rotation around an axis perpendicular to the substrate. Since these degrees of freedom are controlled by friction, predicting the trajectory traced by the origami sheet requires determining the frictional forces. To this end, we first introduce an equivalent spherical linkage to derive explicit equations describing the kinematics of the origami sheet as a function of the folding angle $\beta$ \cite{denavit1955kinematic}. We then numerically identify the three vertices in contact with the substrate upon actuation (see Sections 3A, 3B, and 3C of the Supporting Information for details). Once the contact points and folded configurations are determined, we impose equilibrium to calculate the normal forces at the contact points, $N$, and then compute the static friction forces at each contact point as:

\begin{equation}
\label{eq. 15}
F^s = \mu^s(\psi) N,
\end{equation}

where $\mu^s$ is the experimentally measured static coefficient of friction, which depends on the angle formed by the face of the origami in contact and the substrate, $\psi$ (see Sections 2B).

As an example, in Fig.~\ref{figure_2}A-B, we show the predicted folded configurations, contact points, and evolution of frictional forces for Design III. Our model, consistent with our experiments, predicts a shift in contact points. Specifically, for $\beta \geq 155^\circ$, the model predicts that one edge ($v_3$-$v_6$) and one corner ($v_8$) will be in contact with the substrate, whereas for $\beta < 155^\circ$, there are three point contacts ($v_6$, $v_8$, $v_9$). According to our model, for $\beta \geq 155^\circ$, the edge $v_3$-$v_6$ experiences higher static friction forces, while for $\beta < 155^\circ$, the contact point $v_6$ bears a greater frictional force (Fig.~\ref{figure_2}B).

\begin{figure}[t!]
    \centering
    \includegraphics[width=\linewidth]{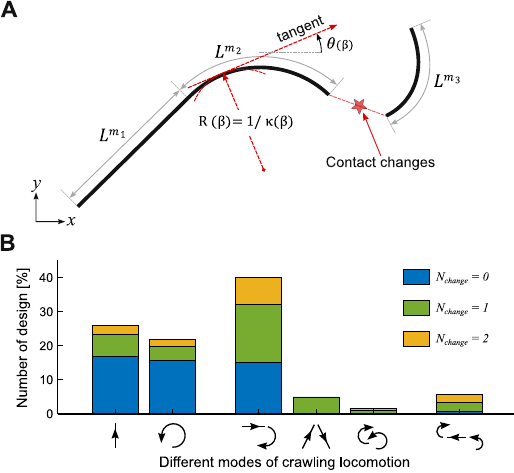} \caption{\textbf{Parametric study.} (A) Schematic of a typical trajectory for the center of mass of an origami sheet upon a loading cycle. (B) Classification of the considered 229,073 designs based on their supported crawling modes. Bar colors indicate the number of
contact changes, $N_\text{change}$.}
    \label{figure_3}
\end{figure}

\begin{figure*}[]
    \centering
    \includegraphics[width=\linewidth]{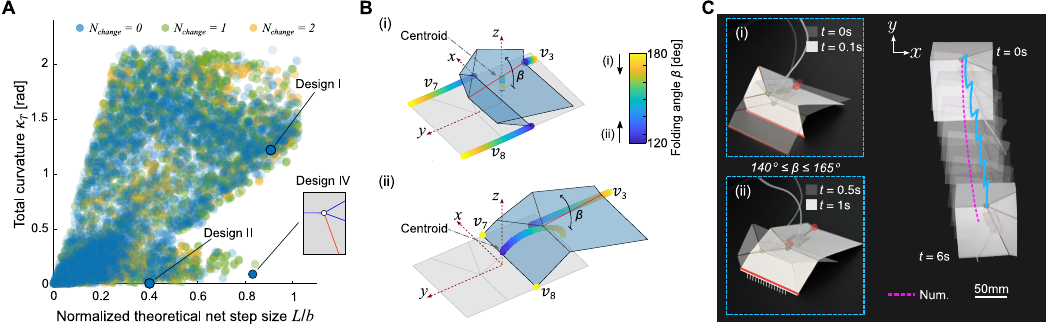} \caption{\textbf{Designs supporting a single mode of crawling.} (A)  Normalized net step size, $L/b$, versus total curvature $\kappa_T$ (rad) for all designs supporting a single mode of crawling.  Marker colors indicate the number of contact changes through a loading cycle, $N_\text{change}$  (B) Numerically predicted evolution of the position of the contact point during (i) inflation and (ii) deflation for Design IV.  (C) Snapshots of Design IV at different times during a loading cycle and frames from videos recorded during tests in which the sample is subjected to multiple inflation/deflation cycles. The magenta dashed line indicates the numerically predicted trajectory.}
    \label{figure_4}
\end{figure*}

Having determined the static frictional forces at the contact points, we proceed to calculate the tangential forces, $\mathbf{f}^{t}$  at these points. Although the system is statically indeterminate (with three equilibrium conditions to determine six unknown tangential force components - $f^{t}{x}$ and $f^{t}{y}$ at each of the three contact nodes), it is crucial to recognize that to achieve locomotion, one contact point must stick while the other two must slip. The sticking contact point must satisfy the sticking criterion:
\begin{equation}
    \label{eq. 17}    
    \mid {f}^{t} \mid \leq \mu^s(\psi) N,  \ \ \ \  
\end{equation}  
whereas the slipping contact points must satisfy:
\begin{equation}
    \label{eq. 18}    
    \mid {f}^{t} \mid = \mu^s(\psi) N.  \\ 
\end{equation} 
Therefore, an origami sheet can crawl upon actuation if the equilibrium equations are satisfied with two contact points meeting Eq.~(\ref{eq. 18}) and one satisfying Eq.~(\ref{eq. 17}). The latter acts as an anchor point, and we fix its in-plane coordinates to calculate the trajectory traced by the sheet as the pouch is deflated (see Section 3D of the Supporting Information for details).

In Figs.~\ref{figure_2}C-D, we present the predicted trajectory traced for Design III for a single cycle of actuation with $157^\circ \leq \beta \leq 163^\circ$ and $130^\circ \leq \beta \leq 135^\circ$, respectively. We observe that in the first case, the sheet moves almost straight, with the tangent to the curve changing by $0.56^\circ$ over an $11.95$ mm step size, while in the second case, the tangent to the curve changes by $1.5^\circ$ over only a $2.14$ mm step size, initiating a turning movement. Upon cyclic actuation, the first scenario results in a straight trajectory, whereas the second scenario leads to a turning motion (see Fig. \ref{figure_2}E). Both outcomes align well with the experimental results, with minor discrepancies attributed to uncertainties in measuring the folding angle, slippage of the anchor point during deflation, and not accounting for centroid displacement during inflation.

Having established a model that accurately captures the trajectory of our origami sheets upon actuation, we then use it to systematically explore the design space. To this end, we investigate the response of more than two million sheets with  $b=90$ mm,  $h \in [1, 1.5]b$,  $\theta_i \in [20^\circ, 30^\circ, 40^\circ, \ldots, 170^\circ]$,  $\theta_v \in [10^\circ, 30^\circ, 50^\circ, \ldots, 350^\circ]$, $x_v \in [-0.7, -0.5, -0.3]b$ and $y_v \in [-1.2, -1, -0.8, -0.6, -0.4]b$. In our analyses, the folding angle $\beta$ is initially decreased from 180$^\circ$ to 120$^\circ$ and then returned to 180$^\circ$ in increments of 0.5$^\circ$. Out of the $579,215$ designs that are rigidly foldable, we find that $243,067$ fail to achieve locomotion due to either the inability to stand on a surface, facet collisions during folding, or the absence of an anchor point during deflation. To rationalize the behavior of the remaining $336,148$ designs that do crawl, we calculate the local curvature 
$\kappa$ of the predicted trajectory during deflation. We classify a portion of the trajectory as straight if $\kappa < 5 \times 10^{-3}$ mm$^{-1}$, and as curved otherwise, using the angle formed by the tangent to the trajectory with the horizontal axis to distinguish between left and right turns (see Fig.~\ref{figure_3}A and Section 3F of the Supporting Information for details).

As shown in Fig. \ref{figure_3}B, we find that many designs support multiple distinct crawling modes. Some achieve multimodal crawling without altering their anchor point ($N_\text{change} = 0$), solely leveraging their nonlinear kinematics. In contrast, others utilize changes in contact points to vary their crawling mode (note that designs with $N_\text{change} > 2$ are disregarded as they typically result in inefficient locomotion). The former results in a continuous trajectory characterized by a broad range of $\kappa$, while the latter exhibits discontinuous transition points along the trajectory (Figure \ref{figure_3}A). Among the 229,073 designs with $N_\text{change} < 2$, $47.77\%$ can support only one crawling mode. Specifically, $25.89\%$ move exclusively straight, and $21.88\%$ turn exclusively. Furthermore, $46.52\%$ of the designs support two modes of crawling: $40 \%$ move both straight and turn, $4.94 \%$ can turn left and right, and $1.58\%$ move straight but can change direction. Finally, $5.71\%$ of the designs exhibit three modes of crawling. By controlling the range of actuation of the folding angle, these designs can crawl straight, turn left, or turn right.

\begin{figure*}[]
    \centering
    \includegraphics[width=\linewidth]{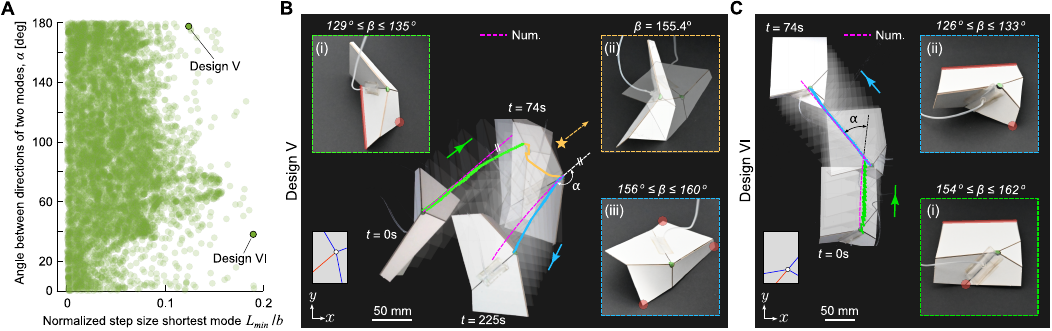} \caption{\textbf{Designs capable of moving straight in two different directions.} (A) Angle between the two supported directions ($\alpha$) versus normalized minimum step size ($L_{min}/b$) for all designs capable of moving straight in two different directions. (B)-(C) Snapshots of (B) Design V and (C) Design VI at different times during a loading cycle, with frames from videos recorded during tests in which the samples are subjected to multiple inflation/deflation cycles. Magenta dashed lines indicate the numerically predicted trajectory.}
    \label{figure_5}
\end{figure*}

\begin{figure*}[]
    \centering
    \includegraphics[width=\linewidth]{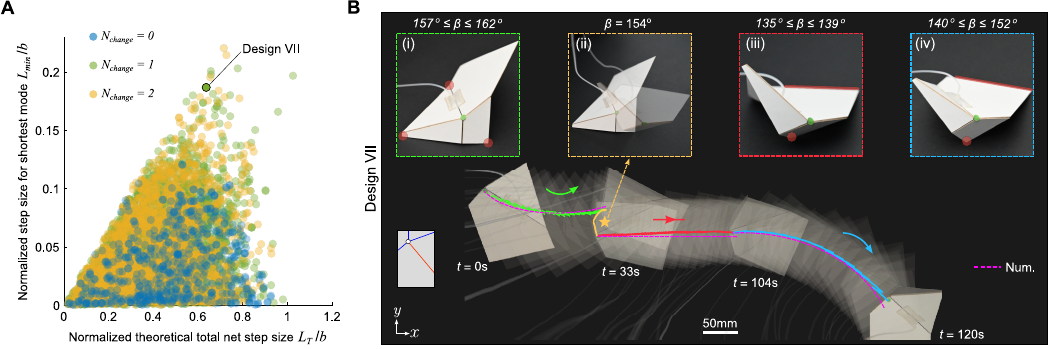} \caption{\textbf{Designs supporting three modes of crawling.} (A) Normalized minimum step size $L_{min}/b$  versus normalized total step size $L_{T}/b$ for all designs supporting three modes of locomotion. Marker colors indicate the number of contact changes through a loading cycle, $N_\text{change}$ (B) Snapshots of Design vII at different times during a loading cycle and frames from videos recorded during tests in which the sample is subjected to multiple inflation/deflation cycles. The magenta dashed line indicates the numerically predicted trajectory.}
    \label{figure_6}
\end{figure*}

In Fig.~\ref{figure_4}, we focus on designs that support only one crawling mode, most of which have $N_\text{change}=0$. To evaluate the performance of these designs, we calculate the length of the trajectory they trace ($L$) and the total change in the angle of the tangent to the trajectory, which is equivalent to the total curvature ($\kappa_T$), as $\beta$ changes from $120^\circ$ to $180^\circ$. In Fig.~\ref{figure_4}A, we report $L/b$ and $\kappa_T$ for all designs. The results indicate that both $L$ and $\kappa_T$ can change largely by tuning the geometry. Specifically, focusing on designs capable of crawling straight (i.e., designs with $\kappa_T \approx 0$), we find that the step size can be greatly increased compared to that of Design II reported in Fig.~\ref{figure_1}. Our model predicts that a design with  $h=1.5b$, ($\theta_1$, $\theta_2$, $\theta_3$, $\theta_v$) = ($150^\circ, 110^\circ, 50^\circ, 30^\circ$) and $(x_v, y_v) = (0.5, 1)b$ (which we refer to as Design IV) will effectively behave as a biped (Fig.~\ref{figure_4}B), achieving $L=0.83b$ (more than twice that of Design II) and $\kappa_T = 0.1$ (see Movie S5). In Fig.~\ref{figure_4}C, we show the experimental results for Design IV, which fully confirmed the numerical predictions in terms of both folding mode and trajectory. In our experiments, we pressurize the pouch attached to the valley fold to vary $\beta$ between $140^\circ$ and   $165^\circ$. In each actuation cycle, the origami sheet moves by approximately $37.2$ mm, which closely matches the model prediction of $42.1$ mm. Consequently, this design requires only six actuation cycles to move approximately $225$ mm, whereas Design II takes fifteen cycles to cover the same distance.

Our systematic exploration of the design space reveals that an origami sheet can not only move straight in one direction or turn but also support two different locomotion modes. These designs include those that can move both straight and turn (similar to Design III in Fig.~\ref{figure_1}E), as well as designs that can either turn or move straight in two different directions. 
In Fig.~\ref{figure_5}, we focus on designs that move straight with a step length $L^{m_1}$ along a direction forming an angle $\alpha^{m_1}$ with the horizontal axis when $\beta_{min}^{m_1}<\beta<\beta_{max}^{m_1}$, and along a direction forming an angle $\alpha^{m_2}$ with steps of length $L^{m_2}$ when $\beta_{min}^{m_2}<\beta<\beta_{max}^{m_2}$. Our model identifies 11,312 such designs, all with $N_\text{change}=1$, and in Fig.~\ref{figure_5}A, we show $\alpha=|\alpha^{m_1}-\alpha^{m_2}|$ and $L_{min}/b=\min(L^{m_1}, L^{m_2})/b$ for all of them. We find that the geometry of the sheets significantly affects both the step size and the angle between the two supported directions of motion. For example, for a design with $h = 1.5b$, $(\theta_1, \theta_2, \theta_3, \theta_v) = (100^\circ, 100^\circ, 60^\circ, 20^\circ)$, and $(x_v, y_v) = (0.7, 0.9)b$ (which we refer to as Design V), our model predicts $\alpha = 177^\circ$ and $L_{min} = 0.12b$. In contrast, for a design with $h = 1.5b$, $(\theta_1, \theta_2, \theta_3, \theta_v) = (140^\circ, 40^\circ, 100^\circ, 50^\circ)$, and $(x_v, y_v) = (0.7, 0.4)b$ (which we refer to as Design VI), it predicts $\alpha = 38^\circ$ and $L_{min} = 0.19b$. Note that for the former ($\beta_{min}^{m1}$, $\beta_{max}^{m1}$, $\beta_{min}^{m2}$, $\beta_{max}^{m2}$)=$(125^\circ, 140 ^\circ, 156^\circ, 165^\circ)$, while for the latter ($\beta_{min}^{m1}$, $\beta_{max}^{m1}$, $\beta_{min}^{m2}$, $\beta_{max}^{m2}$)=$(150 ^\circ, 170^\circ, 120^\circ, 149^\circ)$.

In Figs.~\ref{figure_5}B and~\ref{figure_5}C, we report experimental results for Design V and Design VI, respectively. For Design V, we find that when \(129^\circ \leq \beta \leq 135^\circ\), one edge and a corner are in contact with the substrate (highlighted in red in Fig.~\ref{figure_5}B(i)). However, if \(\beta > 155.4^\circ\), the center of mass of the origami sheet falls outside the triangle formed by the contact points, causing it to tip over (Fig.~\ref{figure_5}B(ii)) and make contact with the substrate through three vertices (highlighted in red in Fig.~\ref{figure_5}B(iii)). In contrast, for Design VI, one edge and a corner are always in contact with the substrate, but the corner in contact changes at \(\beta \approx 149^\circ\) (highlighted in red in Figs.~\ref{figure_5}C(i)-(ii)). These changes in contact make Design V and Design VI move along two straight lines separated by an angle \(\alpha = 165^\circ\) and \(39^\circ\), respectively, when actuated within \(\beta \in [\beta_{min}^{m1}, \beta_{max}^{m1}]\) and \(\beta \in [\beta_{min}^{m2}, \beta_{max}^{m2}]\), see Movie S6 and Movie S7. Thus, our experimental results confirm an effective change in the direction of motion by controlling the range of actuation for $\beta$.

Finally, in Fig.~\ref{figure_6}, we consider designs that can move straight and turn in two different directions. These designs move straight with a step length $L^{m_1}$ when $\beta_{min}^{m_1} < \beta < \beta_{max}^{m_1}$ and turn left and right  with a step length $L^{m_2}$ and $L^{m_3}$ when $\beta_{min}^{m_2} < \beta < \beta_{max}^{m_2}$ and $\beta_{min}^{m_3} < \beta < \beta_{max}^{m_3}$, respectively. Our model identifies 13,070 such designs, with 1,458  achieving this with $N_\text{change}=0$, 5,913  with $N_\text{change}=1$, and 5,699 with $N_\text{change}=2$. In Fig.~\ref{figure_6}A, we report $L_{\text{min}} = \min(L^{m_1}, L^{m_2}, L^{m_3})$ and $L_{T} = \sum_{i=1}^{3} L^{m_i}$ for all these designs. We find that the best-performing designs, characterized by large values for $L_T$ and $L_{\text{min}}$ to ensure large steps when moving both straight and turning in either direction, exploit changes in contact (i.e., they are characterized by $N_\text{change}\geq 1$).
To demonstrate the capability of these designs, we consider an origami sheet with $h = 1.5b$, $(\theta_1, \theta_2, \theta_3, \theta_v) = (130^\circ, 90^\circ, 70^\circ, 90^\circ)$, and $(x_v, y_v) = (0.3, 1.2)b$ (referred to as Design VII). Our model predicts that this design, with $N_\text{change}=1$, has $L_T = 0.65b$ and $L_{\text{min}} = 0.17b$, indicating it can move with large steps both straight and when turning in either direction. In excellent agreement with the model, our experiments show that for $157^\circ \leq \beta \leq 162^\circ$, three vertices are in contact with the substrate (highlighted in red in Fig.~\ref{figure_6}B(i)), resulting in a left-turn crawling motion (green portion of the trajectory in Fig.~\ref{figure_6}B). However, at $\beta \leq 154^\circ$, this contact configuration becomes unstable, and the sheet establishes new contact with the substrate through an edge and a vertex (highlighted in red in Fig.~\ref{figure_6}B(iii)). This change in contact configuration allows Design VII to crawl in a straight line when actuated within the range $135^\circ \leq \beta \leq 139^\circ$ for 67 cycles. Furthermore, when the range of actuation is increased to $140^\circ \leq \beta \leq 152^\circ$ (Fig.~\ref{figure_6}B(iv)), a right-turn crawling motion is achieved with the same contact configuration due to the nonlinear kinematics of the origami (See Movie S8). As such, these results confirm that both changes in contact and nonlinear kinematics can be exploited to control the crawling direction of origami sheets. 

\section*{Discussion and Conclusion}

In summary, we have demonstrated that the nonlinear kinematics of origami sheets can be exploited to realize a crawler capable of multimodal locomotion when actuated with a single input. We specifically considered one of the simplest origami building blocks -- a rigid degree-four vertex -- and identified designs that can move straight and turn both left and right by simply controlling the range of folding for the actuated fold. Future studies could explore an inverse design approach, where an origami crawler is developed to follow a target trajectory. 

While this study focused on crawling, the rich nonlinear mechanics of origami could be further exploited to achieve other locomotion gaits, such as walking, swimming, rolling, and jumping. For instance, transitioning to a circular or curved boundary could enable rolling locomotion by continuously relocating the center of mass \cite{sobolev2023solid}. Additionally, it has been shown that a simple degree-four vertex can be programmed to be multistable \cite{waitukaitis2015origami}. The rapid movements triggered by snapping in multistable designs could potentially be used to make the origami sheets jump \cite{gorissen2020inflatable}. Moreover, multistability could allow for hysteresis-dependent path generation, creating asymmetrical ellipsoidal gaits, and opening up possibilities for walking and swimming locomotion gaits.

In this study, we used an inflatable air pouch for actuation, which limited the folding range from $180^\circ$ to $120^\circ$ and the actuation speed to 1 Hz. By leveraging recent developments in smart and active soft materials, liquid crystal elastomers \cite{kotikian2019untethered} magnetic elastomer \cite{cui2019nanomagnetic} and dielectric elastomer  \cite{sun2022origami} could be integrated into the designs to actuate the folds. Such actuators would make the systems untethered and could enhance their performance and versatility, paving the way for more complex robotic applications.

\section*{Materials and Methods}
\small{Details of the design, materials, and fabrication methods are summarized in Sections S1, S2, and S3. The fabrication process is detailed in Section S1. The experimental procedures, including actuation procedures, testing for the static coefficient of friction, and the pressure-volume relationship of our inflatable pouch motor, are described in Section S2. The numerical model to predict the locomotion behavior of a generic degree-four origami vertex is detailed in Section S3. This includes the geometric description, crawling locomotion analysis, comparison of numerical and experimental results, and the results from the parametric study.}

\section*{Acknowledgments}
\small{\textbf{Funding:} D.F. acknowledges funding support from the Dutch Research Council under the NWO-ENW Rubicon grant agreement (019.191EN.022, 2019). The authors gratefully acknowledge support from the ARO MURI program (W911NF-22-1-0219) and the Simons Collaboration on Extreme Wave Phenomena Based on Symmetries. We thank Bradley But and Anne Meeussen for helpful discussions on the experimental setup.

\section*{Competing interests} 
\small{The authors declare no competing interest.}

\section*{References}
\bibliography{main}

\end{document}